\documentclass[runningheads]{llncs}
\usepackage{graphicx}
\usepackage{comment}
\usepackage{amsmath,amssymb} 
\usepackage{color}

\usepackage{booktabs}
\usepackage{amsfonts}
\usepackage{mathrsfs}
\usepackage{multirow}
\usepackage{color}
\usepackage{subfigure}
\usepackage{colortbl}
\usepackage{tabu}
\usepackage{bbm}
\usepackage{appendix}

\begin{document}
\pagestyle{headings}
\mainmatter
\def\ECCVSubNumber{2748}  

\title{Prototype Rectification for Few-Shot Learning} 

\titlerunning{Prototype Rectification for Few-Shot Learning}
%
\author{Jinlu Liu \and
Liang Song \and
Yongqiang Qin\thanks{Corresponding author.}}
\authorrunning{J. Liu et al.}
%
\institute{AInnovation Technology Co., Ltd.\\
\email{liujinlu, songliang, qinyongqiang@ainnovation.com}}
\maketitle

\begin{abstract}
Few-shot learning requires to recognize novel classes with scarce labeled data. Prototypical network is useful in existing researches, however, training on narrow-size distribution of scarce data usually tends to get biased prototypes. In this paper, we figure out two key influencing factors of the process: \textit{the intra-class bias} and \textit{the cross-class bias}. We then propose a simple yet effective approach for prototype rectification in transductive setting. The approach utilizes label propagation to diminish the intra-class bias and feature shifting to diminish the cross-class bias. We also conduct theoretical analysis to derive its rationality as well as the lower bound of the performance. Effectiveness is shown on three few-shot benchmarks. Notably, our approach achieves state-of-the-art performance on both miniImageNet (70.31\% on 1-shot and 81.89\% on 5-shot) and tieredImageNet (78.74\% on 1-shot and 86.92\% on 5-shot).
\keywords{Few-Shot Learning $\cdot$ Prototype Rectification $\cdot$ Intra-Class Bias $\cdot$ Cross-Class Bias}
\end{abstract}

\section{Introduction}
Many deep learning based methods have achieved significant performance on object recognition tasks with abundant labeled data provided \cite{krizhevsky2012imagenet,russakovsky2015imagenet,he2016deep}. However, these methods generally perform unsatisfactorily if the labeled data is scarce. To reduce the dependency of data annotation, more researchers make efforts to develop powerful methods to learn new concepts from very few samples, which is so-called Few-Shot Learning (FSL) \cite{miller2000learning,fei-fei2006one,vinyals2016matching}. In FSL, we aim to learn prior knowledge on base classes with large amounts of labeled data and utilize the knowledge to recognize few-shot classes with scarce labeled data. It is usually formed as $N$-way $K$-shot few-shot tasks where each task consists of $N$ few-shot classes with $K$ labeled samples per class (the support set) and some unlabeled samples (the query set) for test. 

Classifying test samples by matching them to the nearest class prototype \cite{snell2017prototypical} is a common practice in FSL.
It is supposed that an expected prototype has the minimal distance to all samples within the same class. However, the prototypes they get are always biased due to the data scarcity in few-shot scenarios. The internal factors that restrict the representation ability of the prototypes should be identified for performance improvement. Hence, we figure out the bias in prototype computation and accordingly propose the diminishing methods for rectification.

\begin{figure*}
\centering
\includegraphics[height=2in, width=4.6in]{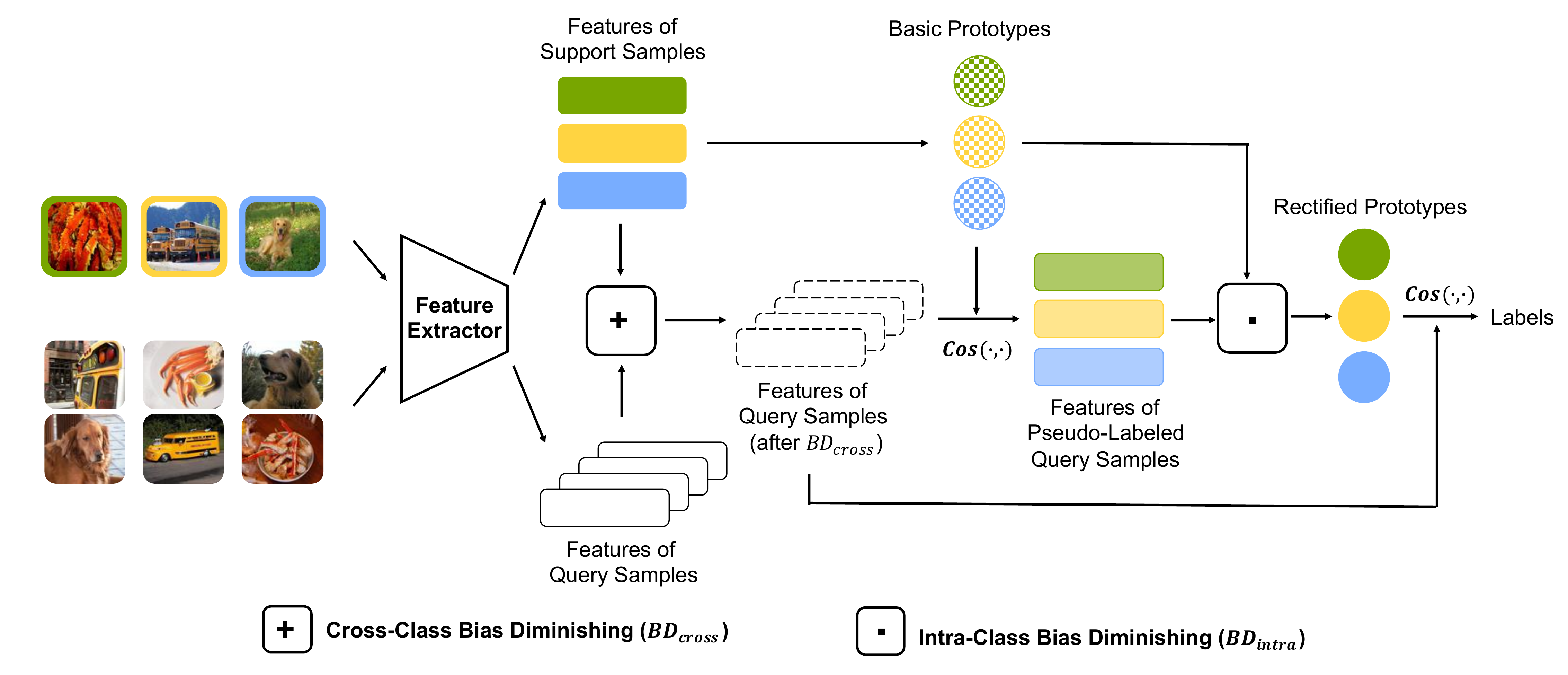}
\caption{Framework of our proposed method for prototype rectification. \textit{The cross-class bias diminishing module} reduces the bias between the support set and the query set while \textit{the intra-class bias diminishing module} reduces the bias between the actually computed prototypes and the expected prototypes.}
\label{Framework}
\end{figure*}

In this paper, we target to find the expected prototypes which have the maximum cosine similarity to all data points within the same class. The cosine similarity based prototypical network (CSPN) is firstly proposed to extract discriminative features and compute basic prototypes from the limited samples. In CSPN, we firstly train a feature extractor with a cosine classifier on the base classes. The cosine classifier has a strong capability of driving the feature extractor to learn discriminative features. It learns an embedding space where features belonging to the same class cluster more tightly. At the inference stage, we use class means as the basic prototypes of few-shot classes. Classification can be directly performed by nearest prototype matching based on cosine similarity.

Since the basic prototypes are biased due to data scarcity, we import a bias diminishing module into the network for prototype rectification, which is called BD-CSPN in this paper. 
We figure out two key factors: the intra-class bias and the cross-class bias, which influence the representativeness of class prototypes. The approach to reduce the bias is accordingly proposed as shown in Fig. \ref{Framework}. The \textit{intra-class bias} refers to the distance between the expectedly unbiased prototype and the prototype actually computed from the available data. To reduce it, we adopt the pseudo-labeling strategy to add unlabeled samples with high prediction confidence into the support set in transductive setting. Considering that some of the pseudo-labeled samples are possibly misclassified, we use the weighted sum as the modified prototypes instead of simple averaging. It avoids bringing larger bias into prototype computation. The \textit{cross-class bias} refers to the distance between the representative vectors of training and test datasets, which are commonly represented as the mean vectors. We reduce it by importing a shifting term $\xi$ to the query samples, driving them to distribute closely to the support samples. 

To verify the rationality of our bias diminishing method, we give the theoretical analysis in Section \ref{theory-analysis}. The derivation of the expected performance of cosine-similarity based prototypical network is firstly given. It shows that the lower bound of the expected accuracy is positively correlated with the number of samples. \textit{We demonstrate the effectiveness and simplicity of our pseudo-labeling strategy in raising the lower bound, which leads to significant improvement as shown in experiments.} Then we give the derivation of shifting term $\xi$ in cross-class bias diminishing.
In conclusion, we argue that our method is simpler yet more efficient than many complicated few-shot learning methods. Also, it is mathematically rigorous with the theoretical analysis. 

Our contributions are summarized as:
\begin{itemize}
    \item[1)] We figure out the internal factors: the intra-class bias and the cross-class bias which restrict the representational ability of class prototypes in few-shot learning.
    \item[2)] We propose the bias diminishing module for prototype rectification, which is mainly conducted by pseudo-labeling and feature shifting. It is conceptually simple but practically effective to improve the performance.
    \item[3)] To verify the rationality of the intra-class bias diminishing method, we theoretically analyze the correlation between the number of sample and the lower bound of the expected performance. Furthermore, we give the derivation of the shifting term in cross-class bias diminishing.
    \item[4)] We conduct extensive experiments on three popular few-shot benchmarks and achieve the state-of-the-art performance. The experiment results demonstrate that our proposed bias diminishing module can bring in significant improvement by a large margin.
\end{itemize}

\section{Related Works}
\textbf{Few-Shot Learning}
Few-shot learning methods can be divided into two groups: \textit{gradient based methods} and \textit{metric learning based methods}. 
Gradient based methods focus on fast adapting model parameters to new tasks through gradient descent \cite{andrychowicz2016learning,finn2017model,li2017meta,nichol2018on,rusu2019meta,lee2019meta}. Typical methods such as MAML \cite{finn2017model} and Reptile \cite{nichol2018on} aim to learn a good parameter initialization that enables the model easy to fine-tune. 
In this section, we focus on metric learning based methods which are more closely to our approach. Metric learning based methods learn an informative metric to indicate the similarity relationship in the embedding space \cite{vinyals2016matching,snell2017prototypical,sung2018learning,allen2019infinite}. 
Relation network \cite{sung2018learning} learns a distance metric to construct the relation of samples within an episode. The unlabeled samples thus can be classified according to the computed relation scores. Prototypical Networks (PN) \cite{snell2017prototypical} views the mean feature as the class prototype and assigns the points to the nearest class prototype based on Euclidean distance in the embedding space. It is indicated in \cite{lee2019meta} that PN shows limited performance in the high-dimensional embedding space.
In some recent works, models trained with a cosine-similarity based classifier are more effective in learning discriminative features \cite{gidaris2018dynamic,chen2019a}. 
In this paper, we use cosine classifier to learn a discriminative embedding space and compute the cosine distance to the class prototype (mean) for classification. The prototype computed in the discriminative feature space is more robust to represent a class.

According to the test setting, FSL can be divided into two branches: \textit{inductive few-shot learning} and \textit{transductive few-shot learning}. The former predicts the test samples one by one while the latter predicts the test samples as a whole. Early proven in \cite{joachims1999transductive,zhou2003learning}, transductive inference outperforms inductive inference especially when training data is scarce. Some literatures recently attack few-shot learning problem in transductive setting.
In \cite{nichol2018on}, the shared information between test samples via normalization is used to improve classification accuracy. Different from \cite{nichol2018on}, TPN \cite{liu2019learning} adopts transductive inference to alleviate low-data problem in few-shot learning. It constructs a graph using the union of the support set and the query set, where labels are propagated from support to query. Under transductive inference, the edge-labeling graph neural network (EGNN) proposed in \cite{kim2019edge} learns more accurate edge-labels through exploring the intra-cluster similarity and the inter-cluster dissimilarity.
Our method takes the advantage of transductive inference that samples with higher prediction confidence can be obtained when the test samples are predicted as a whole.

\textbf{Semi-Supervised Few-Shot Learning}
In semi-supervised few-shot learning, an extra unlabeled set not contained in current episode is used to improve classification accuracy \cite{satorras2018few,ren2018meta,li2019learning}. In \cite{ren2018meta}, the extended versions of Prototypical Networks \cite{snell2017prototypical} are proposed to use unlabeled data to create class prototypes by Soft \textit{k}-Means. LST \cite{li2019learning} employs pseudo-labeling strategy to the unlabeled set, then it re-trains and fine-tunes the base model based on the pseudo-labeled data. For recognizing the novel classes, it utilizes dynamically sampled data which is not contained in the current episode. 
Different from these methods, the unlabeled data in our method comes from the query set and we requires no extra datasets besides the support and query set.

\section{Methodology} 
We firstly use cosine similarity based prototypical network (CSPN) to learn a discriminative feature space and get the basic prototypes of few-shot classes.
Then we figure out two influencing factors in prototype computation: the intra-class bias and the cross-class bias. Accordingly, we propose the bias diminishing (BD) method for prototype rectification in transductive setting.

\subsection{Denotation}  
At the training stage, a labeled dataset $\mathcal{D}$ of base classes $\mathcal{C}_{base}$ is given to train the feature extractor $F_\theta(\cdot)$ and the cosine classifier $C(\cdot|W)$.
At the inference stage, we aim to recognize few-shot classes $\mathcal{C}_{few}$ with $K$ labeled images per class. Episodic sampling is adopted to form such $N$-way $K$-shot tasks. Each episode consists of a support set $\mathcal{S}$ and a query set $\mathcal{Q}$. In the support set, all samples $x$ are labeled and we use the extracted features $X=F_\theta(x)$ to compute the prototypes $P$ of few-shot classes. The samples in the query set are unlabeled for test.

\subsection{Cosine Similarity Based Prototypical Network}
\label{section-CSPN}
We propose a metric learning based method: cosine similarity based prototypical network (CSPN) to compute the basic prototypes of few-shot classes. Training a good feature extractor that can extract discriminative features is of great importance. Thus, we firstly train a feature extractor $F_\theta(\cdot)$ with a cosine similarity based classifier $C(\cdot |W)$ on the base classes. The cosine classifier $C(\cdot |W)$ is:
\begin{equation}
    C(F_\theta (x)\mid W) = Softmax(\tau \cdot Cos(F_\theta (x), W))
\end{equation}
where $W$ is the learnable weight of the base classes and $\tau$ is a scalar parameter. We target to minimize the negative log-likelihood loss on the supervised classification task:
\begin{equation}
    L(\theta, W \mid \mathcal{D}) = \mathbb{E} [-log C(F_\theta (x)\mid W) ]
\end{equation}
At the inference stage, retraining $F_\theta(\cdot)$ and classification weights on the scarce data of $\mathcal{C}_{few}$ classes is likely to run into overfitting. To avoid it, we directly compute the basic prototype $P_n$ of class $n$ as follows:
\begin{equation}
    P_n = \frac{1}{K} {\sum}^K_{i=1}{\overline{X}}_{i,n}
    \label{basic-prototypes}
\end{equation}
where $\overline{X}$ is the normalized feature of support samples. The query samples can be classified by finding the nearest prototype based on cosine similarity.

\subsection{Bias Diminishing for Prototype Rectification}
\label{subsec:bias-diminishing}
In CSPN, we can obtain the basic prototypes by simply averaging the features of support samples. However, the prototypes computed in such low-data regimes are biased against the expected prototypes we want to find. 
Therefore, we identify two influencing factors: the intra-class bias and the cross-class bias, and accordingly propose the bias diminishing approach.

\textbf{The intra-class bias} within a class is defined by Eq. (\ref{intra-bias-equation}):
\begin{equation}
    B_{intra} = {\mathbb{E}}_{X^{'} \sim p_{X^{'}}}[X^{'}] - {\mathbb{E}}_{X \sim p_X}[X]
\label{intra-bias-equation}
\end{equation}
where $p_{X^{'}}$ is the distribution of all data belonging to a certain class and $p_X$ is the distribution of the available labeled data of this class. It is easy to observe the difference between the expectations of the two distributions. The difference becomes more significant in low-data regimes. Since the prototype is computed by feature averaging, the intra-class bias also can be understood as the difference between the expected prototype and the actually computed prototype. The expected prototype is supposed to be represented by the mean feature of all samples within a class. In practice, only a part of samples are available for training which is to say that, it is almost impossible to get the expected prototype. In few-shot  scenario, we merely have $K$ samples per few-shot class. The number of available samples are far less than the expected amount. Computed from scarce samples, the prototypes obviously tend to be biased.

To reduce the bias, we adopt the pseudo-labeling strategy to augment the support set, which assigns temporary labels to the unlabeled data according to their prediction confidence \cite{li2019learning}. Pseudo-labeled samples can be augmented into the support set such that we can compute new prototypes in a `higher-data' regime.
We can simply select top $Z$ confidently predicted query samples per class to augment the support set $\mathcal{S}$ with their pseudo labels. We use CSPN as recognition model to get the prediction scores. Then we have an augmented support set with confidently predicted query samples: $\mathcal{S}'=\mathcal{S} \cup {\mathcal{Q}}^Z_{pseudo}$.
Since some pseudo-labeled samples are likely to be misclassified, simple averaging with the same weights is possible to result in larger bias in prototype computation. To compute the new prototypes in a more reasonable way, we use the weighted sum of $X^{'}$ as the rectified prototype.
We note that $X^{'}$ refers to the feature of the sample in $\mathcal{S}'$ including both original support samples and pseudo-labeled query samples. The rectified prototype of a class is thus computed from the normalized features $\overline X^{'}$:
\begin{equation}
    P^{'}_n = {\sum}^{Z+K}_{i=1} w_{i,n} \cdot {\overline{X}}^{'}_{i,n}
    \label{weighted-sum-prototype}
\end{equation}
where $w_{i,n}$ is the weight indicating the relation of the augmented support samples and the basic prototypes. The weight is computed by:
\begin{equation}
    w_{i,n} = \frac{exp(\varepsilon \cdot Cos(X^{'}_{i,n} ,P_n))}{{\sum}^{K+Z}_{j=1}  exp(\varepsilon \cdot Cos(X^{'}_{j,n} ,P_n))}
    \label{weight}
\end{equation}
$\varepsilon$ is a scalar parameter and $P_n$ is the basic prototype obtained in Section \ref{section-CSPN}. Samples with larger cosine similarity to the basic prototypes hold larger proportions in prototype rectification. Compared with the basic prototype $P_n$, the rectified prototype $P^{'}_n$ distributes closer to the expected prototype.

\textbf{The cross-class bias} refers to the distance between the mean vectors of support and query datasets. It is derived from the domain adaptation problem where the mean value is used as a type of the first order statistic information to represent a dataset \cite{wang2017deep}. Minimizing the distance between different domains is a typical method of mitigating domain gaps. Since the support set and the query set are assumed to distribute in the same domain, the distance between them is the distribution bias rather than the domain gap. The cross-class bias $B_{cross}$ is formulated as:
\begin{equation}
    B_{cross} =  \mathbb{E}_{{X}_s \sim p_{\mathcal{S}}}[{X}_s] - \mathbb{E}_{{X}_q \sim p_{\mathcal{Q}}}[{X}_q]
\label{B_cross}
\end{equation}
where $p_{\mathcal{S}}$ and $p_{\mathcal{Q}}$ respectively represent the distributions of support and query sets. Notably, the support set $\mathcal{S}$ and the query set $\mathcal{Q}$ include $N$ few-shot classes in Eq. (\ref{B_cross}). 
To diminish $B_{cross}$, we can shift the query set towards the support set. In practice, we add a shifting term $\xi$ to each normalized query feature ${\overline{X}}_q$ and $\xi$ is defined as:
\begin{equation}
    \xi = \frac{1}{|\mathcal{S}|}{\sum}^{|\mathcal{S}|}_{i=1}{\overline{X}}_{i,s} - \frac{1}{|\mathcal{Q}|}{\sum}^{|\mathcal{Q}|}_{j=1}{\overline{X}}_{j,q}
\label{cross-bias-xi}
\end{equation}
The detailed derivation of $\xi$ is given in the next section.

\section{Theoretical Analysis}
\label{theory-analysis}
We give the theoretical analysis to show the rationality of our proposed bias diminishing method.
\subsection{Lower Bound of the Expected Performance}
We derive the formulation of our expected performance in theory and point out what factors influence the final result. 
We use $X$ to represent the feature of a class. For clear illustration, the formulation of class prototype we use in this section is given: 
\begin{equation}
    P = \frac{\sum_i^T {\overline{X}}^{'}_i}{T}  
    \label{mean-prototype}
\end{equation}
where $T=K+Z$, ${\overline{X}}^{'}_i \in \mathcal{S}^{'}$ and $\mathcal{S}^{'}$ is a subset sampled from $X$. $\overline{X}$ is the normalized feature and $\overline{P}$ is the normalized prototype.
For cosine similarity based prototypical network, an expected prototype should have the largest cosine similarity to all samples within its class. Our objective is to maximize the expected cosine similarity which is positively correlated with the classification accuracy. It is formulated as:
\begin{equation}
    \max \mathbb{E}_P [\mathbb{E}_X [Cos(P, X)]]
\end{equation}
And we derive it as:
\begin{equation}
\begin{split}
	\mathbb{E}_P [\mathbb{E}_X [Cos(P, X)]] & = \mathbb{E}_{P,X}[\overline P \cdot \overline X] \\
	& = \mathbb{E}[\overline X] \cdot \mathbb{E}[\frac{P}{\|P\|_2}]
\end{split}
\label{objective-derivation}
\end{equation}
From previous works \cite{nowozin2014optimal,rice2015the}, we know that:
\begin{equation}
    \mathbb{E}[\frac{A}{B}]= \frac{\mathbb{E}[A]}{\mathbb{E}[B]} + O(n^{-1}) \quad (first\ order)
    \label{first-order}
\end{equation}
where $A$ and $B$ are random variables. In Eq. (\ref{first-order}), $\frac{\mathbb{E}[A]}{\mathbb{E}[B]}$ is the first order estimator of $\mathbb{E}[\frac{A}{B}]$. Thus, Eq. (\ref{objective-derivation}) is approximate to:
\begin{equation}
    \mathbb{E}_P [\mathbb{E}_X [Cos(P, X)]] \approx \frac{\mathbb{E}[\overline{X}]\cdot \mathbb{E}[P]}{\mathbb{E}[\|P\|_2]}
\end{equation}
Based on Cauchy-Schwarz inequality, we have: 
\begin{equation}
    \mathbb{E}[\|P\|_2] \leq \sqrt{\mathbb{E}[\|P\|_2^2]}    
\end{equation}
$P$ and $\overline X$ are $D$-dimensional vectors which can be denoted as $P=[p_1, p_2,..., p_D]$ and $\overline X = [\overline x_1, \overline x_2,..., \overline x_D]$ respectively. In our method, we assume that each dimension of a vector is independent from each other. Then, we can derive that:
\begin{equation}
\begin{split}
    \mathbb{E}[\|P\|_2^2] = \mathbb{E} [{\sum}^D_{i=1} p^2_i ]& = {\sum}^D_{i=1} [Var[p_i] + \mathbb{E}[p_i]^2] \\
    &= {\sum}^D_{i=1} [Var[p_i] + \mathbb{E}[\overline x_i]^2 ]\\
    &= {\sum}^D_{i=1} [\frac{1}{T} Var[\overline x_i] + \mathbb{E}[\overline x_i]^2 ]
\end{split}
\end{equation}
Thus, the lower bound of the expected cosine similarity is formulated as:
\begin{equation}
\begin{split}
    \mathbb{E}_P [\mathbb{E}_X [Cos(P, X)]] & \geq \frac{\mathbb{E}[\overline{X}]\cdot \mathbb{E}[P]}{\sqrt{\mathbb{E}[\|P\|_2^2]}} \\
    & = \frac{{\sum}^D_{i=1} \mathbb{E}[\overline x_i]^2} {\sqrt{ \frac{1}{T}  {\sum}^D_{i=1}Var[\overline x_i] +  {\sum}^D_{i=1} \mathbb{E}[\overline x_i]^2}}
\end{split}
\label{objective-rewrite}
\end{equation}

Maximizing the expected accuracy is approximate to maximize its lower bound of the cosine similarity as shown in Eq. (\ref{objective-rewrite}). It can be seen that the number $T$ of the sample is positively correlated with the lower bound of the expected performance. Thus, we import more pseudo-labeled samples into prototype computation. \textit{The rationality of the pseudo-labeling strategy in improving few-shot accuracy is that, it can effectively raise the lower bound of the expected performance.}

\subsection{Derivation of Shifting Term $\xi$}
We propose to reduce the cross-class bias by feature shifting and the derivation of shifting term $\xi$ is provided as follows. 
In $N$-way $K$-shot $Q$-query tasks, the accuracy can be formalized as:
\begin{align}
Acc&=\frac{1}{NQ} \sum^N_i \sum_q^Q \mathbbm{1}(y_{i,q}==i)\\
&=\frac{1}{NQ} \sum^N_i  \sum_q^Q \mathbbm{1}(Cos(P_i, X_{i,q})>\max_{j\neq i} \{Cos(P_j, X_{i,q})\})
\label{eqn:acc}
\end{align}
where $y_{i,q}$ is the predicted label and $i$ is the true class label.
$\mathbbm{1}(b)$ is an indicator function. $\mathbbm{1}(b)=1$ if $b$ is true and 0 otherwise. $P_i$ is the prototype of class $i$ and $X_{i,q}$ is the $q$-th query feature of class $i$.
Based on Eq. (\ref{eqn:acc}), the accuracy formulation can be further rewritten as:
\begin{align}
Acc=\frac{1}{NQ} \sum^N_i \sum^Q_q \mathbbm{1}(Cos(P_i, X_{i,q})>t_i)
\end{align}
where $t_i$ denotes the cosine similarity threshold of the $i$-th class. Improving the accuracy is equal to maximize the cosine similarity $Cos(\cdot)$.

As mentioned above, there is a bias between the support and query set of a class $i$. We assume that the bias can be diminished by adding a shifting term $\xi_i$ to the query samples. Since the class labels are unknown, we approximately add the same term $\xi$ to all query samples. The term $\xi$ should follow the objective: 
\begin{align}
\arg\max_\xi \frac{1}{NQ} \sum^N_i \sum^Q_q Cos(P_i, X_{i, q}+\xi)
\label{eqn:intra_obj}
\end{align}

We assume that each feature $X$ can be represented as $X=P+\epsilon$. Eq. (\ref{eqn:intra_obj}) can be further formalized as:
\begin{align}
\arg\max_\xi  \frac{1}{NQ} \sum^N_i \sum^Q_q  Cos(P_i, P_i+\epsilon_{i, q}+\xi)
\end{align}
To maximize the cosine similarity, we should minimize the following objective:
\begin{align}
\min  \frac{1}{NQ} \sum^N_i \sum^Q_q  (\epsilon_{i, q}+\xi)
\end{align}
The term $\xi$ is thus computed:
\begin{align}
\xi&=-E[\epsilon] \\
&= \frac{1}{NQ} \sum_i^N \sum_q^Q (P_i-X_{i, q})
\label{eq-xi}
\end{align}
We can see that Eq. (\ref{eq-xi}) is in line with Eq. (\ref{cross-bias-xi}). For cosine similarity computation, the shifting term is calculated from the normalized features as displayed in Section \ref{subsec:bias-diminishing}.

\section{Experiments}
\label{sec:experiment}
\subsection{Datasets}
\textbf{\textit{miniImageNet}} consists of 100 randomly chosen classes from ILSVRC-2012 \cite{russakovsky2015imagenet}. We adopt the split proposed in \cite{ravi2017optimization} where the 100 classes are split into 64 training classes, 16 validation classes and 20 test classes. Each class contains 600 images of size 84 $\times$ 84.
\textbf{\textit{tieredImageNet}} \cite{ren2018meta} is also a derivative of ILSVRC-2012 \cite{russakovsky2015imagenet} containing 608 low-level categories, which are split into 351, 97, 160 categories for training, validation, test with image size of 84 $\times$ 84.
\textbf{\textit{Meta-Dataset}} \cite{triantafillou2020meta} is a new benchmark that is large-scale and consists of diverse datasets for training and evaluating models.

\subsection{Implementation details}
We train the base recognition model CSPN in the supervised way with SGD optimizer and test the validation set on 5-way 5-shot tasks for model selection. WRN-28-10 \cite{zagoruyko2016wide} is used as the main backbone. ConvNets \cite{gidaris2018dynamic} and ResNet-12 \cite{lee2019meta} are used for ablation. The results are averaged from 600 randomly sampled episodes. Each episode contains 15 query samples per class. The initial value of $\tau$ is 10 and $\varepsilon$ is fixed at 10. More details are shown in the supplementary materials.

\begin{table}
\small
\centering
\caption{Average accuracy (\%) comparison on miniImageNet. $\ddagger$ Training set and validation set are used for training.} 
\begin{tabular}{ccccc}
\hline
\multirow{2}{*}{\textbf{Setting}} &  \multirow{2}{*}{\textbf{Methods}}  & \multirow{2}{*}{\textbf{Backbone}} & \multicolumn{2}{c}{\textbf{miniImageNet}}  \\ 
\cmidrule(r){4-5}
& & & 1-shot & 5-shot \\ \hline
\multirow{12}{*}{Inductive}  & Matching Network \cite{vinyals2016matching} & ConvNet-64 & 43.56$\pm{0.84}$ & 55.31$\pm{0.73}$ \\ 
& MAML \cite{finn2017model} & ConvNet-32 & 48.70$\pm{1.84}$ & 63.11$\pm{0.92}$ \\ 
& Prototypical Networks$\ddagger$ \cite{snell2017prototypical} & ConvNet-64 & 49.42$\pm{0.78}$ & 68.20$\pm{0.66}$   \\
& Relation Net \cite{sung2018learning} &  ConvNet-256 & 50.44$\pm{0.82}$ & 65.32$\pm{0.70}$  \\
& SNAIL \cite{mishra2018a} & ResNet-12 & 55.71$\pm{0.99}$ & 68.88$\pm{0.92}$\\
& LwoF \cite{gidaris2018dynamic} & ConvNet-128 & 56.20$\pm{0.86}$ & 73.00$\pm{0.64}$ \\
& AdaResNet \cite{munkhdalai2018rapid} & ResNet-12 & 56.88$\pm{0.62}$ & 71.94$\pm{0.57}$ \\
& TADAM \cite{oreshkin2018tadam} & ResNet-12 & 58.50$\pm{0.30}$ & 76.70$\pm{0.30}$ \\
& Activation to Parameter$\ddagger$ \cite{qiao2018few} & WRN-28-10 & 59.60$\pm{0.41}$ & 73.74$\pm{0.19}$ \\
& LEO$\ddagger$ \cite{rusu2019meta} & WRN-28-10 & 61.76$\pm{0.08}$ & 77.59$\pm{0.12}$ \\
& MetaOptNet-SVM \cite{lee2019meta} &  ResNet-12 & 62.64$\pm{0.61}$ & 78.63$\pm{0.46}$ \\
& BFSL \cite{gidaris2019boosting} & WRN-28-10 & 62.93$\pm{0.45}$ & 79.87$\pm{0.33}$ \\
\hline
\multirow{2}{*}{Semi-Supervised}  & ML \cite{ren2018meta}  & ConvNet-128 & 49.04$\pm{0.31}$ & 62.96$\pm{0.14}$  \\
& LST \cite{li2019learning} & ResNet-12 & 70.1$\pm{1.9}$ & 78.7$\pm{0.8}$  \\
\hline
\multirow{4}{*}{Transductive} & TPN \cite{liu2019learning} & ConvNet-64 & 55.51$\pm{0.86}$ & 69.86$\pm{0.65}$  \\
& EGNN \cite{kim2019edge} & ConvNet-256 & - & 76.37  \\
& Transductive Fine-Tuning \cite{dhillon2020a} & WRN-28-10 & 65.73$\pm{0.68}$ & 78.40$\pm{0.52}$ \\
& BD-CSPN (ours) & WRN-28-10 & \textbf{70.31$\pm{\textbf{0.93}}$} & \textbf{81.89$\pm{\textbf{0.60}}$}  \\
\hline
\end{tabular} 
\label{fsl-result-miniImagenet}
\end{table}

\begin{table}
\small
\centering
\caption{Average accuracy (\%) comparison on tieredImageNet. * Results by our implementation. $\ddagger$ Training set and validation set are used for training.}
\begin{tabular}{ccccc}
\hline
\multirow{2}{*}{\textbf{Setting}} &  \multirow{2}{*}{\textbf{Methods}}  & \multirow{2}{*}{\textbf{Backbone}} & \multicolumn{2}{c}{\textbf{tieredImageNet}} \\ 
\cmidrule(r){4-5}
& & & 1-shot & 5-shot \\ \hline
\multirow{6}{*}{Inductive}  & MAML \cite{finn2017model} & ConvNet-32 & 51.67$\pm{1.81}$ & 70.30$\pm{1.75}$ \\ 
& Prototypical Networks$\ddagger$ \cite{snell2017prototypical} & ConvNet-64  & 53.31$\pm{0.89}$ & 72.69$\pm{0.74}$  \\
& Relation Net \cite{sung2018learning} &  ConvNet-256  & 54.48$\pm{0.93}$ & 71.32$\pm{0.78}$  \\
& LwoF \cite{gidaris2018dynamic} & ConvNet-128 &  60.35$\pm{0.88}$* & 77.24$\pm{0.72}$* \\
& LEO$\ddagger$ \cite{rusu2019meta} & WRN-28-10 & 66.33$\pm{0.05}$ & 81.44$\pm{0.09}$ \\
& MetaOptNet-SVM \cite{lee2019meta} &  ResNet-12 & 65.99$\pm{0.72}$ & 81.56$\pm{0.53}$ \\
\hline
\multirow{2}{*}{Semi-Supervised} & ML \cite{ren2018meta} & ConvNet-128 & 51.38$\pm{0.38}$ & 69.08$\pm{0.25}$ \\
& LST \cite{li2019learning} & ResNet-12 & 77.7$\pm{1.6}$ & 85.2$\pm{0.8}$ \\
\hline
\multirow{4}{*}{Transductive}  & TPN \cite{liu2019learning} & ConvNet-64 & 59.91$\pm{0.94}$ & 73.30$\pm{0.75}$ \\
& EGNN \cite{kim2019edge} & ConvNet-256 & - & 80.15 \\
& Transductive Fine-Tuning \cite{dhillon2020a} & WRN-28-10 & 73.34$\pm{0.71}$ & 85.50$\pm{0.50}$ \\
& BD-CSPN (ours) & WRN-28-10 & \textbf{78.74$\pm{\textbf{0.95}}$} & \textbf{86.92$\pm{\textbf{0.63}}$} \\
\hline
\end{tabular} 
\label{fsl-result-tieredImagenet}
\end{table}

\subsection{Results on miniImageNet and tieredImageNet}
The results on miniImageNet and tieredImageNet are shown in Table \ref{fsl-result-miniImagenet} and Table \ref{fsl-result-tieredImagenet} respectively. It can be seen that we achieve state-of-the-art performance in all cases.  
Compared with existing transductive methods \cite{liu2019learning,kim2019edge,dhillon2020a}, our proposed BD-CSPN consistently achieves the best performance on both datasets. EGNN \cite{kim2019edge} transductively learns edge-labels through exploring the intra-cluster similarity and the inter-cluster dissimilarity. Transductive Fine-Tuning \cite{dhillon2020a} is newly published, providing a strong baseline by simple fine-tuning techniques.
In comparison with TPN \cite{liu2019learning}, we achieve better results with a simpler implementation of label propagation technique. Given the similar backbone ConvNet-128 on miniImageNet, BD-CSPN produces good results of 61.74\% and 76.12\% on 1-shot and 5-shot tasks respectively, surpassing TPN by large margins.

Our method also shows superiority compared with existing semi-supervised methods \cite{ren2018meta,li2019learning}. Note that LST \cite{li2019learning} uses extra unlabeled data as auxiliary information in evaluation, which is not contained in current episode. It re-trains and fine-tunes the model on each novel task. We have a simpler technique without re-training and fine-tuning which is more efficient in computation. 

\subsection{Results on Meta-Dataset}
To further illustrate the effectiveness of our method, we show 5-shot results on the newly proposed Meta-Dataset \cite{triantafillou2020meta} in Table \ref{meta-datasets}. The average rank of our 5-shot model is \textbf{1.9}. More details are provided in our supplementary materials.
\begin{table}
\centering
\caption{5-shot results on Meta-Dataset: the model is trained on ILSVRC-2012 only and test on the listed test sources. }
\label{meta-datasets}
\resizebox{\textwidth}{6mm}{
\begin{tabular}{rcrcrcrcrc}
\hline
Test &  5-shot & Test &  5-shot & Test &  5-shot & Test &  5-shot & Test &  5-shot \\ 
\hline
 ILSVRC & 59.80 & Omniglot  &78.29  & Aircraft&43.42 & Birds & 67.22 & Textures &54.82  \\
 Quick Draw & 58.80 &  Fungi & 61.56 & VGG Flower & 83.88 & Traffic Signs & 68.68 & MSCOCO & 52.69 \\
\hline

\end{tabular} 
}
\end{table}

\subsection{Ablation Study}
The ablation results are shown in Table \ref{ablation-bias-results}.
We display the results of CSPN as baselines which are obtained in inductive setting. The network is trained on traditional supervised tasks (64-way), following the setting in \cite{gidaris2018dynamic,gidaris2019boosting}. It achieves better performance than some complicated meta-trained methods \cite{qiao2018few,rusu2019meta} with the same backbone, as shown in Table \ref{fsl-result-miniImagenet}. Based on CSPN, our BD module makes an improvement by large margins up to 9\% and 3\% on 1-shot and 5-shot tasks respectively. It leads to relatively minor improvements in 5-shot scenarios.

\begin{table}
\small
\centering
\caption{Ablative results of bias diminishing module. CSPN: without bias diminishing modules; $BD_{c}$-CSPN: with cross-class bias diminishing module; $BD_{i}$-CSPN: with intra-class bias diminishing module; $BD$-CSPN: with both modules.}
\begin{tabular}{cccccccc}
\hline
Dataset &  Method & 1-shot & 5-shot & Dataset &  Method & 1-shot & 5-shot \\ 
\hline
\multirow{4}{*}{miniImageNet} & CSPN & 61.84 & 78.64 & \multirow{4}{*}{tieredImageNet} & CSPN & 69.20 & 84.31  \\
& $BD_{c}$-CSPN & 62.54 & 79.32 & & $BD_{c}$-CSPN & 70.84 & 84.99 \\
& $BD_{i}$-CSPN & 69.81 & 81.58 & & $BD_{i}$-CSPN & 78.12 & 86.67\\
& $BD$-CSPN & \textbf{70.31} & \textbf{81.89} & & $BD$-CSPN & \textbf{78.74} & \textbf{86.92}   \\
\hline
\end{tabular} 
\label{ablation-bias-results}
\end{table}

\subsubsection{Ablation of Intra-Class Bias Diminishing}
It can be seen in Table \ref{ablation-bias-results} that $BD_{i}$-CSPN brings in significant improvements on both datasets. The intra-class bias diminishing module especially shows its merit in 1-shot scenarios. With intra-class bias diminished, the accuracy on 1-shot miniImageNet increases from 61.84\% to 69.81\% and the accuracy on 1-shot tieredImageNet raises to 78.12\% from 69.20\%.

\begin{figure*}
\centering
\subfigure[]{
\begin{minipage}[t]{0.33\linewidth}
\centering
\includegraphics[width=1.55in, height=1.3in]{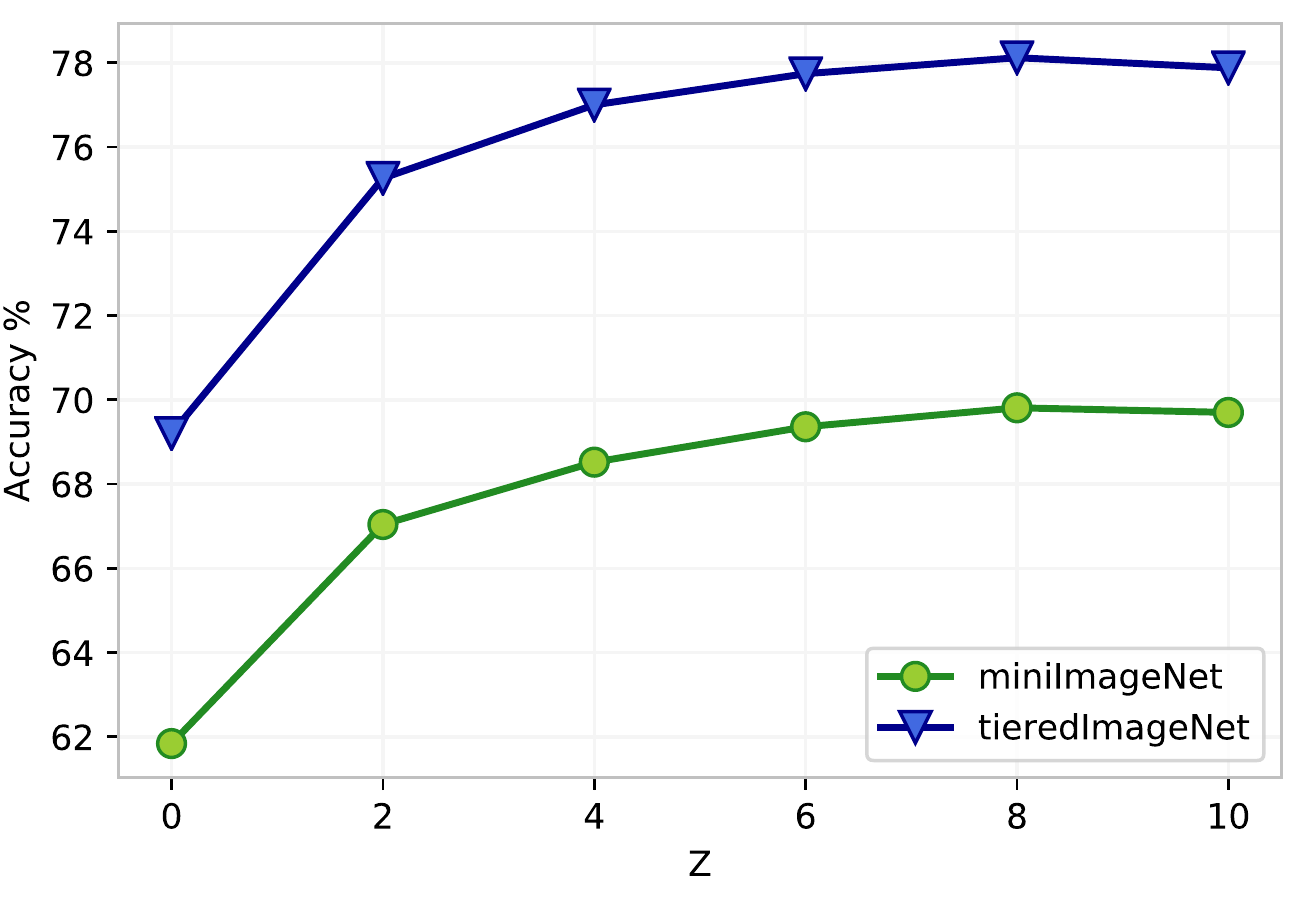}
\label{intra-bias-1-shot}
\end{minipage}%
}%
\subfigure[]{
\begin{minipage}[t]{0.33\linewidth}
\centering
\includegraphics[width=1.55in, height=1.3in]{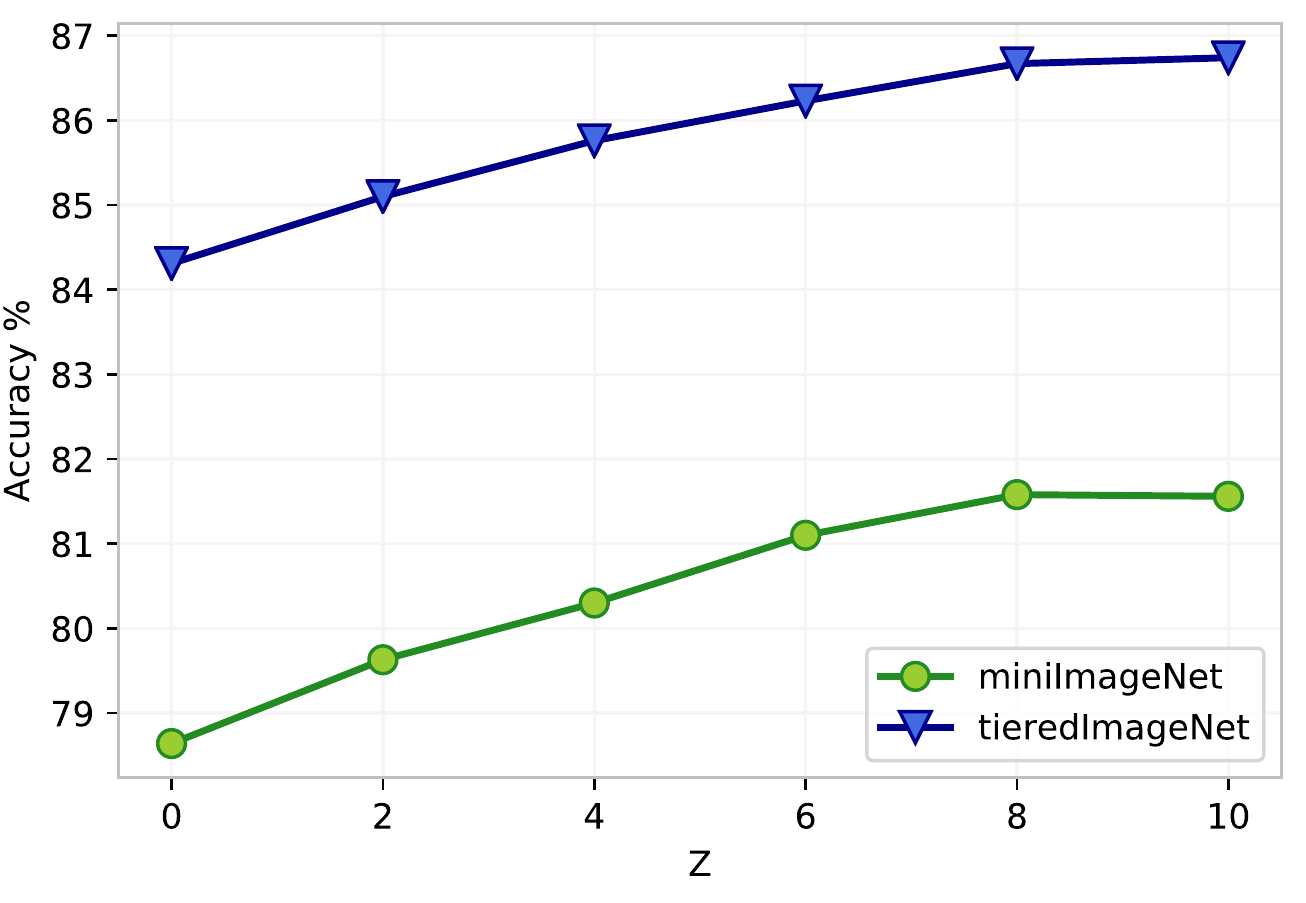}
\label{intra-bias-5-shot}
\end{minipage}%
}%
\subfigure[]{
\begin{minipage}[t]{0.33\linewidth}
\includegraphics[width=1.55in, height=1.3in]{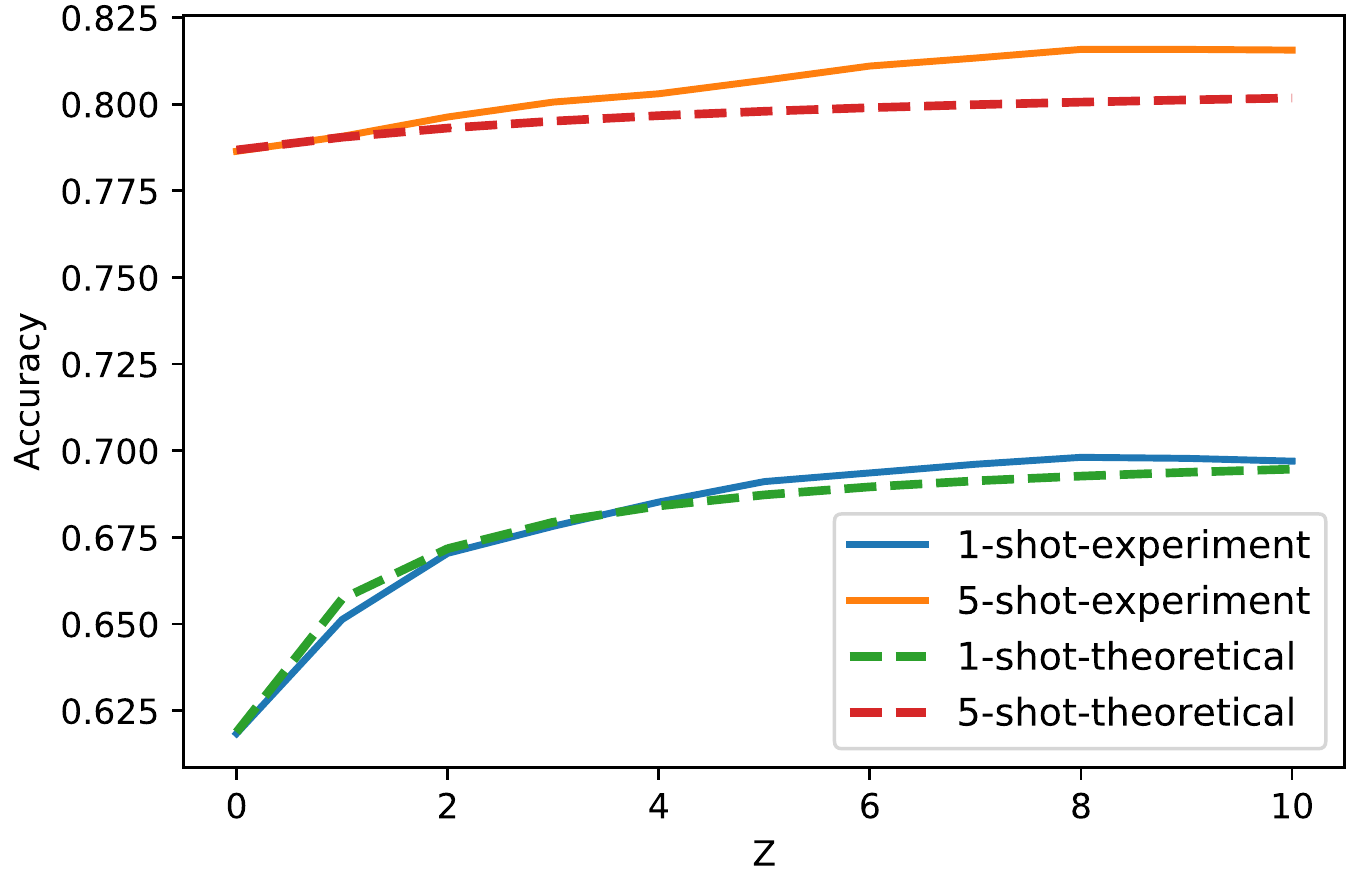}
\label{curve-fitting}
\end{minipage}%
}%
\caption{Effectiveness of intra-bias diminishing. $Z$: the number of pseudo-labeled samples. (a) 5-way 1-shot results. (b) 5-way 5-shot results. (c) Theoretical value on miniImageNet. The experiment results (solid lines) show a consistent tendency with the theoretical results (dashed lines). }
\label{intra-bias-comparison}
\end{figure*}

Furthermore, to intuitively demonstrate the influence of our proposed intra-class bias diminishing module, we display the 5-way accuracy in Fig. \ref{intra-bias-1-shot}-\ref{intra-bias-5-shot}. The results are reported without using cross-class bias diminishing module.
It shows a coincident tendency that with more pseudo-labeled samples, there is an obvious growth of classification accuracy. We use the validation set to determine the value of $Z$ and set it to 8 for accuracy comparison in Table \ref{fsl-result-miniImagenet} and Table \ref{fsl-result-tieredImagenet}. 

\textbf{\textit{Theoretical Value}}
As we know, the expected accuracy $Acc(P, X)$ has a positive correlation with the expected cosine similarity. Then we derive the first-order estimation of $Acc(P, X)$ from Eq. (\ref{objective-rewrite}) which is formulated as:
\begin{equation}
\small
    Acc(P, X) \approx \eta \cdot \frac{\alpha}{\sqrt{\lambda \cdot \frac{1}{K+Z} + \alpha}}
    \label{objective-final}
\end{equation}
where $\eta$ is a coefficient and $K+Z=T$. $\lambda$ and $\alpha$ are values correlated with the variance term and the expectation term in Eq. (\ref{objective-rewrite}). The theoretical values of $\lambda$ and $\alpha$ can be approximately computed from the extracted features. Furthermore, we can compute the value of $\eta$ by 1-shot and 5-shot accuracies of CSPN. Thus, the number $Z$ is the only variable in Eq. (\ref{objective-final}). The theoretical curves are displayed as the dashed lines in Fig. \ref{curve-fitting} to show the impact of $Z$ on classification accuracy. The dashed lines, showing the theoretical lower bound of the expected accuracy, have a consistent tendency with our experiment results in Fig. \ref{intra-bias-1-shot}-\ref{intra-bias-5-shot}. Since the cosine similarity is continuous and the accuracy is discrete, the accuracy stops increasing when the cosine similarity grows to a certain value.

\textbf{\textit{T-SNE Visualization}}
We show t-SNE visualization of our intra-bias diminishing method in Fig. \ref{tsne-visualization} for intuitive illustration. The basic prototype of each class is computed from the support set while the rectified prototype is computed from the augmented support set. In this section, the expected prototype refers to the first term in Eq. (\ref{intra-bias-equation}) which is represented by the average vector of all samples (both support and query samples) of a class in an episode. Due to the scarcity of labeled samples, there is a large bias between the basic prototype and the expected prototype. The bias can be reflected by the distance between the stars and the triangles in Fig. \ref{tsne-visualization}. 

\subsubsection{Ablation of Cross-Class Bias Diminishing} 
Table \ref{ablation-bias-results} shows the ablative results of the cross-class bias diminishing module. It illustrates an overall improvement as a result of diminishing the cross-class bias. Moving the whole query set towards the support set center by importing the shifting term $\xi$ is an effective approach to reduce the bias between the two datasets. For example, the accuracy increases by 1.64\% on 1-shot tieredImageNet.

\begin{figure}
\centering
\subfigure[T-SNE visualization of $BD_{intra}$]{
\begin{minipage}[t]{0.5\linewidth}
\centering
\includegraphics[width=1.7in, height=1.55in]{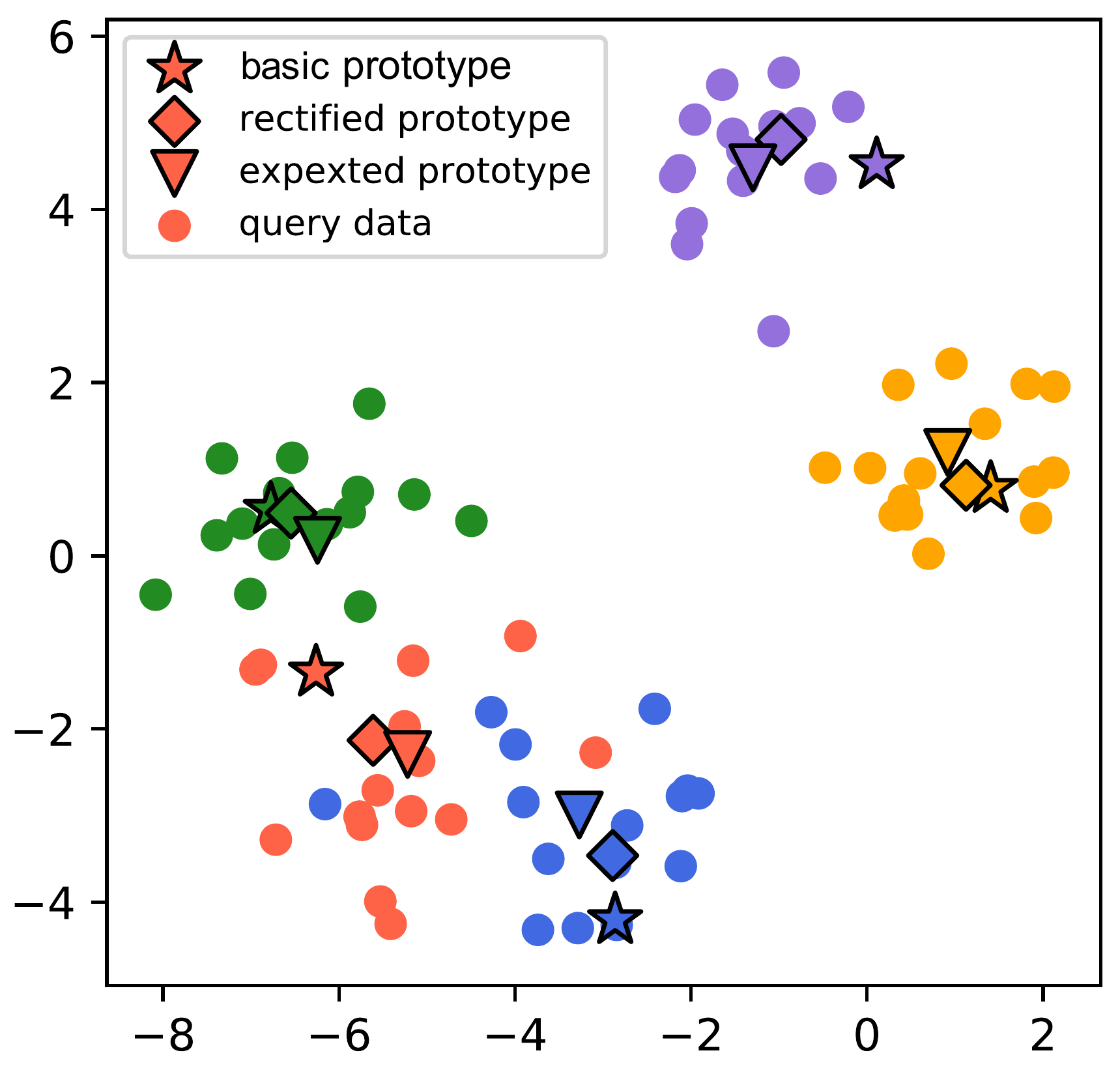}
\label{tsne-visualization}
\end{minipage}%
}%
\subfigure[T-SNE visualization of $BD_{cross}$]{
\begin{minipage}[t]{0.5\linewidth}
\centering
\includegraphics[width=1.7in, height=1.55in]{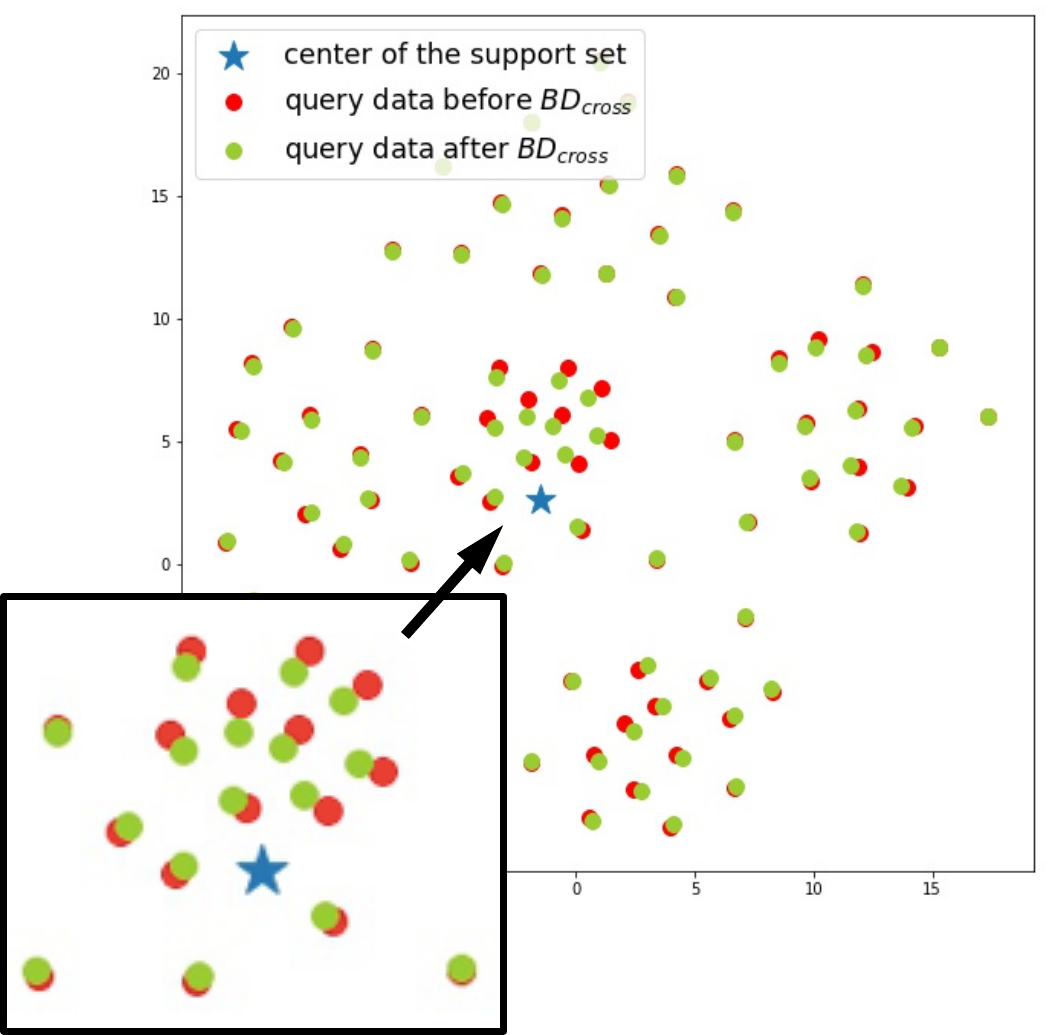}
\label{cross-bias-tsne-visualization}
\end{minipage}%
}%
\centering
\caption{We randomly sample a 5-way 1-shot episode on tieredImageNet. Different classes are marked in different colors. Best viewed in color with zoom in.}
\end{figure}

\textbf{\textit{T-SNE Visualization} }
In few-shot learning, the support set includes far less samples compared with the query set in an episode. There exists a large distance between the two mean vectors of the datasets. We aim to decrease the distance by shifting the query samples towards the center of the support set as shown in Fig. \ref{cross-bias-tsne-visualization}.
It depicts the spatial changing of the query samples, before and after cross-class bias diminishing. The significant part is zoomed in for clear visualization, where the query samples with ${BD}_{cross}$ (marked in green) distribute more closely to the center of support set.

\begin{table}
\small
\centering
\caption{Ablation of \textbf{backbones} and result \textbf{comparison with TFT} (Transductive Fine-Tuning \cite{dhillon2020a}) on miniImageNet. * The backbone is ConvNet-64.}
\begin{tabular}{lccc|lccc}
\hline
1-shot & CSPN & BD-CSPN & TFT & 5-shot & CSPN & BD-CSPN & TFT \\ \hline
 ConvNet-128 & 55.62 & \textbf{61.74} & 50.46* &  ConvNet-128 &  72.57 & \textbf{76.12} & 66.68* \\ 
 ResNet-12 &  59.14 & \textbf{65.94} & 62.35 &  ResNet-12  & 76.26 & \textbf{79.23} & 74.53  \\ 
 WRN-28-10 &  61.84 & \textbf{70.31} & 65.73  & WRN-28-10  & 78.64 & \textbf{81.89}  & 78.40 \\ \hline
\end{tabular}
\label{comparison-pn-cspn}
\end{table}

\subsubsection{Ablation of Backbone}
The results on miniImageNet are displayed in Table \ref{comparison-pn-cspn} and more ablation results are given in the supplementary materials. Our method also shows good performance based on ConvNet-128 and ResNet-12, which is better than most approaches in Table \ref{fsl-result-miniImagenet}. For example, with ResNet-12, we achieve 79.23\% in 5-shot scenario, outperforming the strongest baselines: 78.7\% \cite{li2019learning} and 78.63\% \cite{lee2019meta}.

\subsection{Comparison with Transductive Fine-Tuning}
We compare our method with TFT \cite{dhillon2020a} in Table \ref{comparison-pn-cspn}, which is recently proposed as a new baseline for few-shot image classification. BD-CSPN outperforms it given different backbones. For example, we achieve better results which are higher than TFT by 3\% to 5\% given ResNet-12. Since BD-CSPN and TFT conduct experiments in the same transductive setting, the comparison between these two methods is more persuasive to demonstrate the effectiveness of the approach.

\section{Conclusions}
In this paper, we propose a powerful method of prototype rectification in few-shot learning, which is to diminish the intra-class bias and the cross-class bias of class prototypes. Our theoretical analysis verifies that, the proposed bias diminishing method is effective in raising the lower bound of the expected performance.
Extensive experiments on three few-shot benchmarks demonstrate the effectiveness of our method. The proposed bias diminishing method achieves significant improvements in transductive setting by large margins (e.g. 8.47\% on 1-shot miniImageNet and 9.54\% on 1-shot tieredImageNet).

\appendix

\section{Appendix}

\subsection{Implementation Details}
WRN-28-10 \cite{zagoruyko2016wide}, is used as the main backbone in the experiments. ConvNet-64 \cite{dhillon2020a}, ConvNet-128 \cite{gidaris2018dynamic}, ConvNet-256 \cite{kim2019edge} and ResNet-12 \cite{lee2019meta} are used in ablation study. We remove the last ReLU layer of WRN-28-10 in experiments. The results reported in our experiments are collected by sampling 600 episodes with $95\%$ confidence intervals. We choose SGD as the optimizer with a momentum of 0.9 and a weight decay parameter of 0.0005. The maximum training epoch is set to 60. The initial learning rate is 0.1 and it is reduced after 10, 20, 40 epochs. At the training stage, we use horizontal flip and random crop on the two ImageNet derivatives as in \cite{gidaris2018dynamic,qiao2018few,lee2019meta}.

\subsection{Results on Omniglot and CUB}
We conduct extra experiments on another two benchmarks: Omniglot \cite{lake2011one} and CUB \cite{wah2011the}.

\subsubsection{Omniglot}

\textit{Omniglot} has 1623 classes of handwritten characters with 20 samples per class. All images are resized to 28 x 28. The data augmentation techniques proposed by \cite{santoro2016meta,snell2017prototypical} are used in higher-way test, which rotates each image by 90, 180, 270 degrees to form new classes. Therefore, the dataset has total 6492 classes and we use 4112 classes for training, 688 classes for validation and 1692 classes for test as in \cite{snell2017prototypical}.

\begin{table}
\centering
\caption{Results on Omniglot.}
\begin{tabular}{lcccc}
\toprule
\multicolumn{1}{c}{{ }}& \multicolumn{2}{c}{{ \textbf{1-shot}}} & \multicolumn{2}{c}{{\textbf{5-shot}}} \\
\multicolumn{1}{c}{\multirow{-2}{*}{{ \textbf{Omniglot}}}} & {CSPN}  & {BD-CSPN} & {CSPN}  & {BD-CSPN} \\ \midrule
{ConvNet-64} & {97.40} & {99.62}   & {99.60} & {99.76}   \\
{ConvNet-128}& {97.33} & {99.69}   & {99.63} & {99.75}   \\
{ConvNet-256}  & {97.85} & {99.77}   & {99.63} & {99.76}   \\
{ResNet-12} & {98.70} & {99.80}   & {99.72} & {99.77}   \\
{WRN-28-10} & {99.02} & {99.08}   & {99.82} & {99.85}  \\  \midrule
\end{tabular}
\end{table}

\subsubsection{CUB}

We use the Caltech-UCSD Birds (CUB) 200-2011 dataset \cite{wah2011the} of 200 fine-grained bird species. The dataset is split into 100 training classes, 50 validation classes and 50 test classes as provided in \cite{chen2019a}.

\begin{table}[htbp]
\centering
\caption{Results on CUB.}
\begin{tabular}{lcccc}
\toprule
\multicolumn{1}{c}{}& \multicolumn{2}{c}{{\textbf{1-shot}}} & \multicolumn{2}{c}{{\textbf{5-shot}}} \\ 
\multicolumn{1}{c}{\multirow{-2}{*}{{\textbf{CUB}}}} & {CSPN}  & {BD-CSPN} & {CSPN}  & {BD-CSPN} \\  \midrule
{ConvNet-64}& {64.72} & {75.10}   & {84.21} & {87.25}   \\
{ConvNet-128}& {65.86} & {76.11}   & {85.97} & {87.52}   \\ 
{ConvNet-256}& {65.99} & {75.77}   & {84.74} & {87.76}   \\
{ResNet-12}& {76.24} & {84.90}   & {88.68} & {90.22}   \\ 
{WRN-28-10}& {77.80} & {87.45}   & {90.14} & {91.74}   \\ \midrule
\end{tabular}
\end{table}

\vspace*{2\baselineskip}

\subsection{Additional Ablation on miniImageNet and tieredImageNet}
We provide supplementary ablation study on miniImageNet and tieredImageNet to show our performance on different backbones.

\begin{table*}
\centering
\caption{Backbone ablation on miniImageNet.}
\begin{tabular}{lcc}
\toprule
{\textbf{miniImageNet}} & {\textbf{1-shot}} & {\textbf{5-shot}} \\ \midrule
{ConvNet-64}            & {60.48}           & {75.02}           \\
{ConvNet-256}           & {60.97}           & {75.19}          \\ \midrule
\end{tabular}
\end{table*}

\begin{table}
\centering
\caption{Backbone ablation on tieredImageNet.}
\begin{tabular}{lcc}
\toprule
{\textbf{tieredImageNet}} & {\textbf{1-shot}} & {\textbf{5-shot}} \\ \midrule
{ConvNet-64}              & {65.08}           & {78.08}           \\
{ConvNet-128}             & {66.33}           & {79.57}           \\
{ConvNet-256}             & {67.09}           & {80.66}           \\
{ResNet-12}               & {76.17}           & {85.70}          \\ \midrule
\end{tabular}
\end{table}

\subsection{Higher-way Results}
Results on higher-way tasks are given in Table \ref{higher-mini} to Table \ref{higher-omni} to show the effectiveness of our method in harder tasks.

\vspace*{4\baselineskip}

\begin{table}
\centering
\caption{Higher-way test on miniImageNet.}
\begin{tabular}{lcc}
\toprule
{\textbf{miniImageNet}} & {\textbf{1-shot}} & {\textbf{5-shot}} \\ \midrule
{10-way}                & {51.58}           & {69.35}           \\
{20-way}                & {36.00}           & {55.23}           \\ \midrule
\end{tabular}
\label{higher-mini}
\end{table}

\begin{table}
\centering
\caption{Higher-way test on tieredImageNet.}
\begin{tabular}{lcc}
\toprule
{\textbf{tieredImageNet}} & {\textbf{1-shot}} & {\textbf{5-shot}} \\ \midrule
{10-way}                  & {63.39}           & {77.54}           \\
{20-way}                  & {48.48}           & {65.68}           \\
{50-way}                  & {31.67}           & {49.50}          \\ \midrule
\end{tabular}
\label{higher-tiered}
\end{table}

\begin{table}
\centering
\caption{Higher-way test on Omniglot.}
\begin{tabular}{llcccc}
\toprule
{\textbf{Omniglot}}& & \multicolumn{2}{c}{{\textbf{1shot}}}& \multicolumn{2}{c}{{\textbf{5shot}}} \\ 
& & {CSPN}& {BD-CSPN} & {CSPN}  & {BD-CSPN} \\ \midrule
{10-way}& {ConvNet-128}& {92.83}& {98.46}  & {98.67} & {99.02}  \\
 &{ConvNet-256} & {93.82} & {98.65}  & {98.90} & {99.14}  \\
& {ResNet-12}& {96.38}& {98.97}  & {99.11} & {99.22}  \\
& {WRN-28-10}& {96.62}& {99.12}  & {99.35} & {99.40}  \\
{200-way}& {ConvNet-64}& {75.44}& {89.08}  & {93.21} & {94.72}  \\
{1000-way}& {ConvNet-64}& {56.85}& {71.18}  & {82.72} & {85.87}  \\ \midrule
\end{tabular}
\label{higher-omni}
\end{table}

\subsection{Robust Test}
We conduct an experiment as follows to test the robustness of the proposed BD-CSPN. In each 5-way K-shot 15-query episode, we randomly add extra 15$\times$N' samples of N' classes that do not belong to the 5 classes. The extra samples are treated as unlabeled data. Our model shows good robustness (aka little performance drop) in 5-shot cases. The accuracy decreases to some extents when the unlabeled data increases.

\begin{table}
\centering
\caption{Robust test on miniImageNet. Acc: the accuracy of the labeled 5$\times$15 query data. mAP: it is computed from top-15 confidently predicted data of each class.}
\begin{tabular}{lll}
\toprule
{\textbf{miniImageNet}} & {\textbf{N'=1}}           & {\textbf{N'=5}}           \\ \midrule
{1-shot Acc}            & {66.88 (3.43$\downarrow$)} & {64.58 (5.73$\downarrow$)} \\
{5-shot Acc}            & {80.31 (1.58$\downarrow$)} & {79.25 (2.64$\downarrow$)}  \\ 
1-shot mAP & 76.08 & 64.35  \\
5-shot mAP & 89.03 & 81.06  \\
\midrule
\end{tabular}
\end{table}

\subsection{Results on Meta-Dataset}
Meta-Dataset \cite{triantafillou2020meta} is a new benchmark for few-shot learning. It is large-scale and consists of diverse datasets for training and evaluating models. We show our results in Table \ref{meta-dataset} and the ranks of our 5-shot model. For detailed comparison, please refer to \textit{Table 1 (top) in \cite{triantafillou2020meta}}.

\begin{table}
\centering
\caption{Results on Meta-Dataset. Avg. rank of our 5-shot model is \textbf{1.9}.}
\begin{tabular}{lcc}
\toprule
Test Source & 1-shot & 5-shot  \\ \midrule
{ILSVRC}          & {45.57}           & {59.80 (1)}                            \\
{Omniglot}        & {66.77}           & {78.29 (1)}                            \\
{Aircraft}        & {32.85}           & {43.42 (7)}                            \\
{Birds}           & {49.41}           & {67.22 (3)}                            \\
{Textures}        & {40.64}           & {54.82 (1)}                            \\
{Quick Draw}       & {45.52}           & {58.80 (1)}                            \\
{Fungi}           & {44.65}           & {61.56 (1)}                            \\ 
{VGG Flower}       & {69.97}           & {83.88 (4)}                            \\
{Traffic Signs}    & {53.93}           & {68.68 (1)}                            \\ 
{MSCOCO}          & {40.06}           & {52.69 (1)}                           \\ \midrule
\end{tabular}
\label{meta-dataset}
\end{table}

\bibliographystyle{splncs04}
\bibliography{2748}

\begin{thebibliography}{10}
\providecommand{\url}[1]{\texttt{#1}}
\providecommand{\urlprefix}{URL }
\providecommand{\doi}[1]{https://doi.org/#1}

\bibitem{allen2019infinite}
{Allen}, K., {Shelhamer}, E., {Shin}, H., {Tenenbaum}, J.: Infinite mixture
  prototypes for few-shot learning. In: ICML. pp. 232--241 (2019)

\bibitem{andrychowicz2016learning}
{Andrychowicz}, M., {Denil}, M., {Gomez}, S., {Hoffman}, M.W., {Pfau}, D.,
  {Schaul}, T., {Shillingford}, B., de~{Freitas}, N.: Learning to learn by
  gradient descent by gradient descent. In: NIPS. pp. 3981--3989 (2016)

\bibitem{chen2019a}
{Chen}, W.Y., {Liu}, Y.C., {Kira}, Z., {Wang}, Y.C.F., {Huang}, J.B.: A closer
  look at few-shot classification. In: ICLR (2019)

\bibitem{dhillon2020a}
{Dhillon}, G.S., {Chaudhari}, P., {Ravichandran}, A., {Soatto}, S.: A baseline
  for few-shot image classification. In: ICLR (2020)

\bibitem{fei-fei2006one}
{Fei-Fei}, L., {Fergus}, R., {Perona}, P.: One-shot learning of object
  categories. vol.~28, pp. 594--611 (2006)

\bibitem{finn2017model}
{Finn}, C., {Abbeel}, P., {Levine}, S.: Model-agnostic meta-learning for fast
  adaptation of deep networks. In: ICML. pp. 1126--1135 (2017)

\bibitem{gidaris2019boosting}
{Gidaris}, S., {Bursuc}, A., {Komodakis}, N., {Perez}, P.P., {Cord}, M.:
  Boosting few-shot visual learning with self-supervision. In: ICCV. pp.
  8058--8067 (2019)

\bibitem{gidaris2018dynamic}
{Gidaris}, S., {Komodakis}, N.: Dynamic few-shot visual learning without
  forgetting. In: CVPR. pp. 4367--4375 (2018)

\bibitem{he2016deep}
{He}, K., {Zhang}, X., {Ren}, S., {Sun}, J.: Deep residual learning for image
  recognition. In: CVPR. pp. 770--778 (2016)

\bibitem{joachims1999transductive}
{Joachims}, T.: Transductive inference for text classification using support
  vector machines. In: ICML. pp. 200--209 (1999)

\bibitem{kim2019edge}
{Kim}, J., {Kim}, T., {Kim}, S., {Yoo}, C.D.: Edge-labeling graph neural
  network for few-shot learning. In: CVPR. pp. 11--20 (2019)

\bibitem{krizhevsky2012imagenet}
{Krizhevsky}, A., {Sutskever}, I., {Hinton}, G.E.: Imagenet classification with
  deep convolutional neural networks. vol.~141, pp. 1097--1105 (2012)

\bibitem{lake2011one}
{Lake}, B.M., {Salakhutdinov}, R., {Gross}, J., {Tenenbaum}, J.B.: One shot
  learning of simple visual concepts. Cognitive Science  \textbf{33}(33) (2011)

\bibitem{lee2019meta}
{Lee}, K., {Maji}, S., {Ravichandran}, A., {Soatto}, S.: Meta-learning with
  differentiable convex optimization. In: CVPR. pp. 10657--10665 (2019)

\bibitem{li2019learning}
{Li}, X., {Sun}, Q., {Liu}, Y., {Zhou}, Q., {Zheng}, S., {Chua}, T.S.,
  {Schiele}, B.: Learning to self-train for semi-supervised few-shot
  classification. In: NeurIPS (2019)

\bibitem{li2017meta}
{Li}, Z., {Zhou}, F., {Chen}, F., {Li}, H.: Meta-sgd: Learning to learn quickly
  for few shot learning. (2017)

\bibitem{liu2019learning}
{Liu}, Y., {Lee}, J., {Park}, M., {Kim}, S., {Yang}, E., {Hwang}, S.J., {Yang},
  Y.: Learning to propagate labels: Transductive propagation network for
  few-shot learning. In: ICLR (2019)

\bibitem{miller2000learning}
{Miller}, E., {Matsakis}, N., {Viola}, P.: Learning from one example through
  shared densities on transforms. In: CVPR. vol.~1, pp. 464--471 (2000)

\bibitem{mishra2018a}
{Mishra}, N., {Rohaninejad}, M., {Chen}, X., {Abbeel}, P.: A simple neural
  attentive meta-learner. In: ICLR (2018)

\bibitem{munkhdalai2018rapid}
{Munkhdalai}, T., {Yuan}, X., {Mehri}, S., {Trischler}, A.: Rapid adaptation
  with conditionally shifted neurons. In: ICML. pp. 3661--3670 (2018)

\bibitem{nichol2018on}
{Nichol}, A., {Achiam}, J., {Schulman}, J.: On first-order meta-learning
  algorithms. (2018)

\bibitem{nowozin2014optimal}
{Nowozin}, S.: Optimal decisions from probabilistic models: The
  intersection-over-union case. In: CVPR. pp. 548--555 (2014)

\bibitem{oreshkin2018tadam}
{Oreshkin}, B.N., {López}, P.R., {Lacoste}, A.: Tadam: Task dependent adaptive
  metric for improved few-shot learning. In: NIPS. pp. 721--731 (2018)

\bibitem{qiao2018few}
{Qiao}, S., {Liu}, C., {Shen}, W., {Yuille}, A.L.: Few-shot image recognition
  by predicting parameters from activations. In: CVPR. pp. 7229--7238 (2018)

\bibitem{ravi2017optimization}
{Ravi}, S., {Larochelle}, H.: Optimization as a model for few-shot learning.
  In: ICLR (2017)

\bibitem{ren2018meta}
{Ren}, M., {Ravi}, S., {Triantafillou}, E., {Snell}, J., {Swersky}, K.,
  {Tenenbaum}, J.B., {Larochelle}, H., {Zemel}, R.S.: Meta-learning for
  semi-supervised few-shot classification. In: ICLR (2018)

\bibitem{rice2015the}
Rice, S.H.: The expected value of the ratio of correlated random variables.
  Texas Tech University  (2015)

\bibitem{russakovsky2015imagenet}
{Russakovsky}, O., {Deng}, J., {Su}, H., {Krause}, J., {Satheesh}, S., {Ma},
  S., {Huang}, Z., {Karpathy}, A., {Khosla}, A., {Bernstein}, M.S., {Berg},
  A.C., {Fei-Fei}, L.: Imagenet large scale visual recognition challenge.
  vol.~115, pp. 211--252 (2015)

\bibitem{rusu2019meta}
{Rusu}, A.A., {Rao}, D., {Sygnowski}, J., {Vinyals}, O., {Pascanu}, R.,
  {Osindero}, S., {Hadsell}, R.: Meta-learning with latent embedding
  optimization. In: ICLR (2019)

\bibitem{santoro2016meta}
{Santoro}, A., {Bartunov}, S., {Botvinick}, M., {Wierstra}, D., {Lillicrap},
  T.: Meta-learning with memory-augmented neural networks. In: ICML. pp.
  1842--1850 (2016)

\bibitem{satorras2018few}
{Satorras}, V.G., {Estrach}, J.B.: Few-shot learning with graph neural
  networks. In: ICLR (2018)

\bibitem{snell2017prototypical}
{Snell}, J., {Swersky}, K., {Zemel}, R.S.: Prototypical networks for few-shot
  learning. In: NIPS. pp. 4077--4087 (2017)

\bibitem{sung2018learning}
{Sung}, F., {Yang}, Y., {Zhang}, L., {Xiang}, T., {Torr}, P.H., {Hospedales},
  T.M.: Learning to compare: Relation network for few-shot learning. In: CVPR.
  pp. 1199--1208 (2018)

\bibitem{triantafillou2020meta}
{Triantafillou}, E., {Zhu}, T., {Dumoulin}, V., {Lamblin}, P., {Evci}, U.,
  {Xu}, K., {Goroshin}, R., {Gelada}, C., {Swersky}, K., {Manzagol}, P.A.,
  {Larochelle}, H.: Meta-dataset: A dataset of datasets for learning to learn
  from few examples. In: ICLR (2020)

\bibitem{vinyals2016matching}
{Vinyals}, O., {Blundell}, C., {Lillicrap}, T.P., {Kavukcuoglu}, K.,
  {Wierstra}, D.: Matching networks for one shot learning. In: NIPS. pp.
  3637--3645 (2016)

\bibitem{wah2011the}
{Wah}, C., {Branson}, S., {Welinder}, P., {Perona}, P., {Belongie}, S.: The
  caltech-ucsd birds-200-2011 dataset. California Institute of Technology
  (2011)

\bibitem{wang2017deep}
{Wang}, Y., {Li}, W., {Dai}, D., {Gool}, L.V.: Deep domain adaptation by
  geodesic distance minimization. In: ICCVW. pp. 2651--2657 (2017)

\bibitem{zagoruyko2016wide}
{Zagoruyko}, S., {Komodakis}, N.: Wide residual networks. In: BMVC (2016)

\bibitem{zhou2003learning}
{Zhou}, D., {Bousquet}, O., {Lal}, T.N., {Weston}, J., {Schölkopf}, B.:
  Learning with local and global consistency. In: NIPS. pp. 321--328 (2003)

\end{thebibliography}
\end{document}